%% file: paper.tex
\documentclass[]{bytedance_seed}

\usepackage[toc,page,header]{appendix}

\usepackage{minitoc}
\usepackage{amssymb}

\usepackage{makecell} 
\usepackage{algorithm}      
\usepackage{algpseudocode}  

\usepackage{graphicx}

\usepackage{listings}
\usepackage{url}
\usepackage{booktabs} 
\usepackage{wrapfig} 
\newcommand{\tablestyle}[2]{\setlength{\tabcolsep}{#1}\renewcommand{\arraystretch}{#2}\centering\footnotesize}

\usepackage{colortbl} 
\usepackage{xcolor}

\usepackage{caption}
\usepackage{subcaption}

\def\eg{\emph{e.g.}} 
 
\title{ThinkRL-Edit: Thinking in Reinforcement Learning for\\Reasoning-Centric Image Editing}

\author[1,2,*]{Hengjia Li}
\author[2, \dagger]{Liming Jiang}
\author[2]{Qing Yan}
\author[2]{Yizhi Song}
\author[2]{Hao Kang}
\author[2]{Zichuan Liu}
\author[2]{Xin Lu}
\author[1, \ddagger]{Boxi Wu}
\author[1]{Deng Cai}

\affiliation[1]{Zhejiang University}
\affiliation[2]{Intelligent Creation, ByteDance}

\contribution[*]{Work done during internship at ByteDance}
\contribution[\dagger]{Project Lead}
\contribution[\ddagger]{Corresponding Asuthor}

\abstract{
Instruction-driven image editing with unified multimodal generative models has advanced rapidly, yet their underlying visual reasoning remains limited, leading to suboptimal performance on reasoning-centric edits. Reinforcement learning (RL) has been investigated for improving the quality of image editing, but it faces three key challenges: (1) limited reasoning exploration confined to denoising stochasticity, (2) biased reward fusion, and (3) unstable VLM-based instruction rewards.
In this work, we propose \textbf{ThinkRL-Edit}, a reasoning-centric RL framework that decouples visual reasoning from image synthesis and expands reasoning exploration beyond denoising. To the end, we introduce Chain-of-Thought (CoT)–based reasoning sampling with planning and reflection stages prior to generation in online sampling, compelling the model to explore multiple semantic hypotheses and validate their plausibility before committing to a visual outcome. 
To avoid the failures of weighted aggregation, we propose an unbiased chain preference grouping strategy across multiple reward dimensions. Moreover, we replace interval-based VLM scores with a binary checklist, yielding more precise, lower-variance, and interpretable rewards for complex reasoning. Experiments show our method significantly outperforms prior work on reasoning-centric image editing, producing instruction-faithful, visually coherent, and semantically grounded edits.
}

\date{\today}

\checkdata[Project Page]{\url{https://echopluto.github.io/ThinkRL-page/}}

\begin{document}
\maketitle


\input{sec/1_intro}
\input{sec/2_related}

\input{sec/3_method}

\input{sec/4_exp}

\input{sec/5_con}

\clearpage

\bibliographystyle{plainnat}
\bibliography{main}




\end{document}

%% file: sec/1_intro.tex
\section{Introduction}
\label{sec:intro}

\begin{figure*}[ht]
\centering
\includegraphics[width=\linewidth]{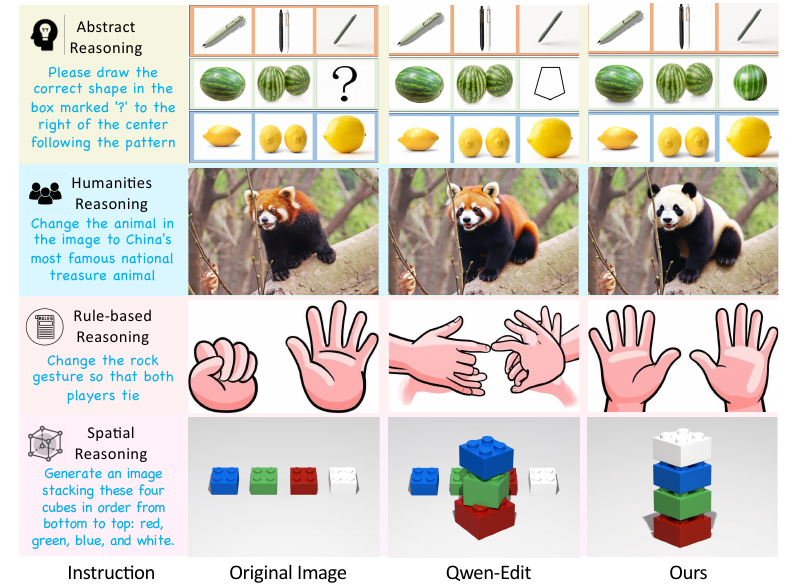}
\vspace{-0.3cm}
\caption{\textbf{Comparisons on reasoning-centric image editing.}
Although unified multimodal generative models such as Qwen-Edit~\cite{qwen-image} have substantially improved editing quality, their underlying reasoning remains underexplored, especially for reasoning-centric editing. In contrast, our method delivers accurate edits with deep reasoning, achieving strong consistency and high perceptual quality across diverse reasoning-driven editing scenarios.}
\label{fig:teaser}
\vspace{-0.4cm}
\end{figure*}

\begin{figure*}[ht]
\centering
\includegraphics[width=.97\linewidth]{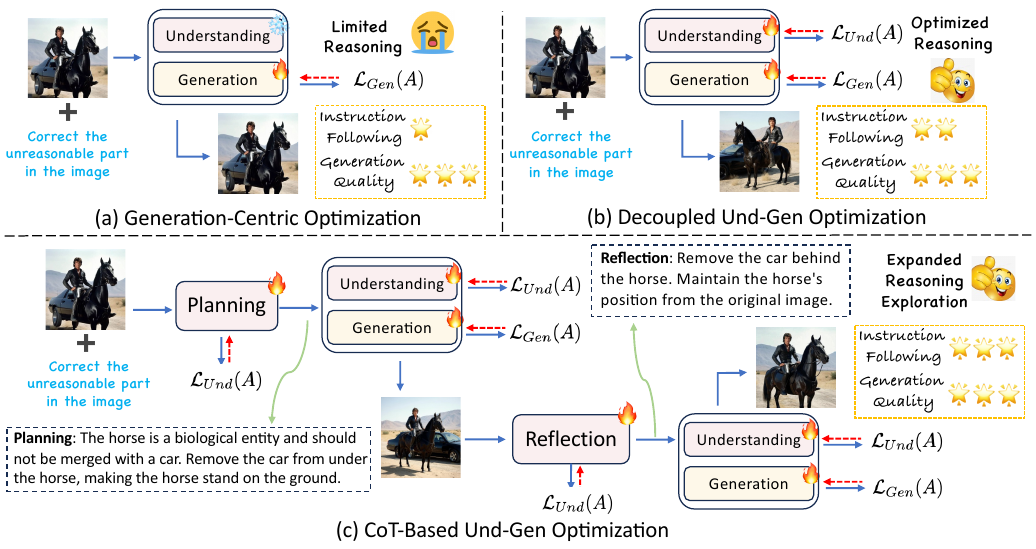}
\vspace{-0.3cm}
\caption{\textbf{Comparison with prior methods.} Prior RL methods for visual generation~\cite{liu2025flow, xue2025dancegrpo} focus on exploration within the stochastic space of generation, improving synthesis quality but offering limited reasoning capability. To address this issue, we decouple and optimize the understanding and generation modules to preserve high-fidelity synthesis while enabling exploration of optimal trajectories in the reasoning space. Besides, we introduce CoT-based sampling and optimization to further expand stochastic exploration over reasoning pathways.}
\label{fig:motivation}
\vspace{-0.4cm}
\end{figure*}

Recent progress in unified multi-modal generative models~\cite{xin2025lumina,wu2025omnigen2,wang2024emu3, wu2025qwen, lin2025uniworld, li2025uniworldv2} has significantly advanced instruction-driven image editing. 
However, despite impressive visual fidelity, the reasoning capability behind such edits remains largely underexplored. In particular, reasoning-centric editing requires models to thoroughly understand both the reference image and the given instruction before synthesis, rather than merely producing visually plausible content as illustrated in \cref{fig:teaser}. 

Prior efforts have explored reinforcement learning (RL)~\cite{shao2024deepseekmath,wei2025skywork,xue2025dancegrpo,liu2025flow, wang2025pref} to substantially improve the editing quality. However, they exhibit clear challenges when applied to reasoning-centric image editing, which requires not only high-fidelity synthesis but also strong visual reasoning prior to generation. Three major challenges arise:
\begin{itemize}
\item \textbf{Limited reasoning exploration.} Existing RL approaches typically restrict exploration to stochasticity within the denoising process while the reasoning processes underpinning the edits remain under-explored as shown in \cref{fig:motivation}. For example, FlowGRPO~\cite{liu2025flow} expands the search space by converting ODE-based denoising into SDE-based sampling, yet it neglects exploration across diverse visual reasoning trajectories. Thus, these methods are better suited for text rendering and aesthetic enhancement, but fundamentally insufficient for reasoning-driven editing, where reasoning must precede generation.

\item \textbf{Biased reward aggregation.} Editing requires balancing instruction fidelity, visual consistency, and generation quality. Previous methods~\cite{xue2025dancegrpo,liu2025flow} typically combine these rewards using simple weighted sums. This naive aggregation is highly vulnerable to edge cases. For example, an unchanged image may obtain a very high consistency score, while an instruction-accurate edit might be unfairly penalized for larger semantic changes.

\item \textbf{Unstable instruction rewards.} Prior works often rely on vision-language models (VLMs)~\cite{Qwen25VL, chen2024internvl} to assign discrete instruction-following scores (e.g., 1–5). However, such reward signals are high-variance and inconsistent, especially for complex reasoning tasks, where repeated evaluations frequently produce differing results.
\end{itemize}
In this work, we address these challenges by introducing a reasoning-centric RL framework for instruction-based image editing that decouples reasoning–generation during exploration. Specifically, to expand the exploration space beyond denoising stochasticity and enable optimization over diverse reasoning trajectories, we explicitly separate and optimize visual reasoning prior to image generation. Furthermore, we introduce chain-of-thought (CoT)~\cite{wei2022chain} sampling, incorporating planning and reflection stages prior to image generation. This design compels the model to explore multiple semantic hypotheses and evaluate their plausibility before committing to a visual outcome. It helps the model establish correct semantic interpretations not just by seeking `better denoising', but by first discovering and refining meaningful visual reasoning paths.

Besides, to avoid the limitations of naive weighted reward fusion, which often collapses towards trivial solutions or overfits individual objectives, we introduce an unbiased chain preference optimization strategy that holistically ranks reasoning chains across all reward dimensions. Instead of collapsing rewards into one scalar, we jointly sort sampled chains per group and update gradients only from chains that form a consistent total order. This captures unified preferences across objectives (e.g., instruction faithfulness, visual coherence, perceptual quality) and prevents trivial solutions or overfitting to single objectives.

Furthermore, to provide more precise and stable reasoning rewards from vision-language models, we replace interval-based scoring with a checklist evaluation. For each editing instruction, we derive binary questions from the reference image and prompt, have the VLM answer yes/no, and use the count of “yes” as the alignment score. Experiments show this fine-grained reasoning reward yields more accurate, lower-variance, and more interpretable rewards, especially for complex reasoning where scalar scores fluctuate or miss nuanced compliance.

Extensive experiments demonstrate that our approach significantly outperforms prior methods on reasoning-centric image editing tasks, producing edits that are not only instruction-faithful but also visually coherent and semantically grounded. In summary, our contributions are as follows.
\begin{itemize}
\item We propose to decouple visual reasoning from image synthesis and further introduce CoT-based reasoning sampling to explore diverse trajectories before generation.

\item We introduce a unbiased ranking-based grouping strategy that orders sampled reasoning chains across multiple reward dimensions, avoiding weighted-fusion collapse.

\item We replace interval-based VLM scoring with a binary checklist from the reference image and instruction, yielding more precise, lower-variance, and interpretable rewards for complex reasoning.

\item We conduct comprehensive experiments across multiple benchmarks, demonstrating that our method substantially outperforms prior works on reasoning-centric image editing.
\end{itemize}


%% file: sec/2_related.tex
\section{Related Work}
\label{sec:formatting}

\subsection{Reasoning-Centric Image Editing}
Reasoning-centric image editing models aim to bridge high-level semantic understanding and reasoning of textual instructions with precise visual manipulation. Traditional approaches achieve accurate editing by modifying the diffusion trajectory without additional training, such as partial denoising from intermediate SDE steps~\cite{meng2021sdedit}, cross-attention control~\cite{hertz2022prompt,wang2023enhancing}, mask-guided blending~\cite{wang2025dreamtext,avrahami2022blended,wang2024primecomposer}, CLIP- or diffusion-guided manipulation~\cite{kim2022diffusionclip, li2024unihda, li2023few, li2024gca}, and latent inversion for fidelity preservation~\cite{wang2024magicface,kawar2023imagic}.
Despite their strong controllability, these methods lack the capacity to handle complex, reasoning-intensive semantic edits.
Recent unified multimodal models advance in a complementary direction by employing a single framework for both image understanding and editing~\cite{xin2025lumina,wu2025omnigen2,wang2024emu3, wu2025qwen, lin2025uniworld, li2025uniworldv2}.
For example, Bagel~\cite{deng2025bagel} introduces a \emph{think} mode that first generates reasoning text to enhance instruction fidelity and semantic consistency during editing.
However, despite these advances, current models still struggle with tasks that require deeper logical reasoning and multi-step inference during visual editing.

\begin{figure*}[t]
\centering
\includegraphics[width=\linewidth]{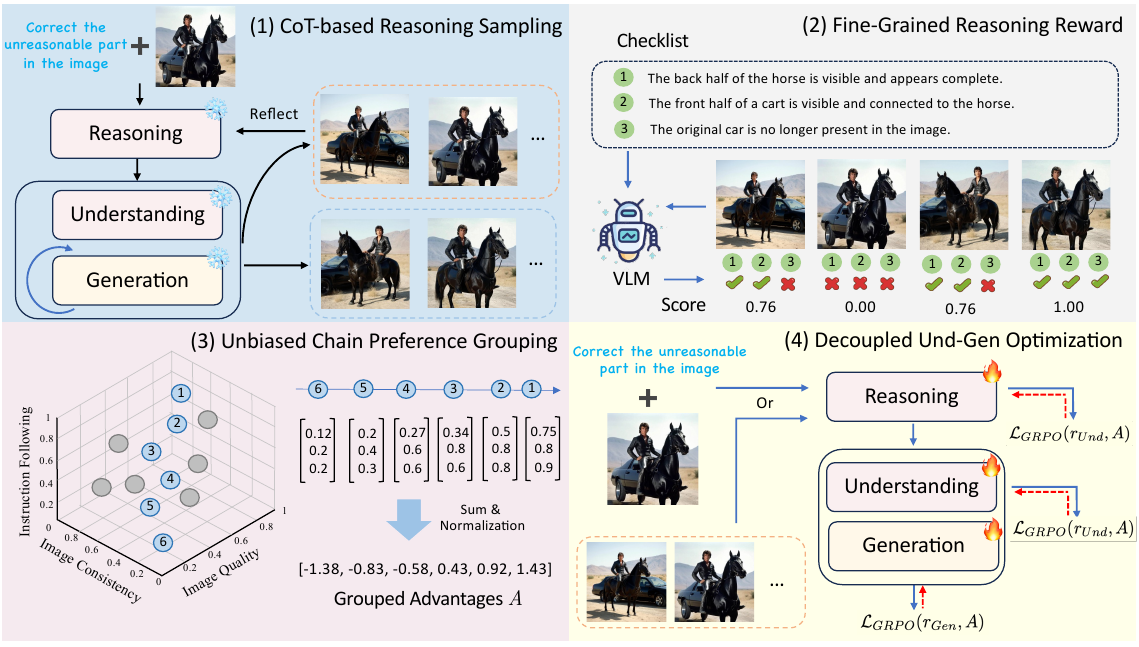}
\vspace{-0.5cm}
\caption{\textbf{Overview of our method.} During sampling, we perform Chain-of-Thought reasoning with explicit planning and reflection to enlarge stochasticity in the reasoning space. For rewards, a fine-grained, sample-specific checklist guides the VLM to produce accurate and stable reasoning scores. In grouping, we construct an unbiased preference chain across candidates to select training samples and compute advantages $A$. Finally, policy updates apply a unified editing reward while decoupling updates to the reasoning, understanding, and generation modules, enhancing reasoning capability without sacrificing synthesis quality.}
\label{fig:method}
\vspace{-0.3cm}
\end{figure*}

\subsection{Reinforcement Learning for Visual Generation}
Reinforcement Learning from Human Feedback (RLHF)~\citep{ouyang2022training} has emerged as the dominant paradigm for aligning large language models (LLMs) to be more helpful~\citep{hu2022lora,shao2024deepseekmath}, honest~\citep{gao2024honestllm}, and harmless~\citep{yang2025asft}.
Inspired by its success in language alignment, recent studies have extended RL-based frameworks to text-to-image (T2I) generation~\citep{black2023training}, typically by training a reward model (RM) on human preference data~\citep{xu2024visionreward} or prompt-image alignment scores~\citep{xu2023imagereward}.
Building on this foundation, advanced algorithms such as Group Relative Policy Optimization (GRPO)~\citep{shao2024deepseekmath,wei2025skywork,xue2025dancegrpo,liu2025flow, wang2025pref} have shown strong potential in aligning both diffusion and flow-matching models.
For example, FlowGRPO reformulates the deterministic ODE process of flow matching into a stochastic differential equation, effectively expanding the exploration space of denoising trajectories.
However, these methods largely overlook the semantic reasoning search space, and their reward models remain limited in evaluating reasoning-intensive editing tasks, resulting in suboptimal performance when complex logical inference is required during visual editing.

\begin{figure*}[h]
\centering
\includegraphics[width=.97\linewidth]{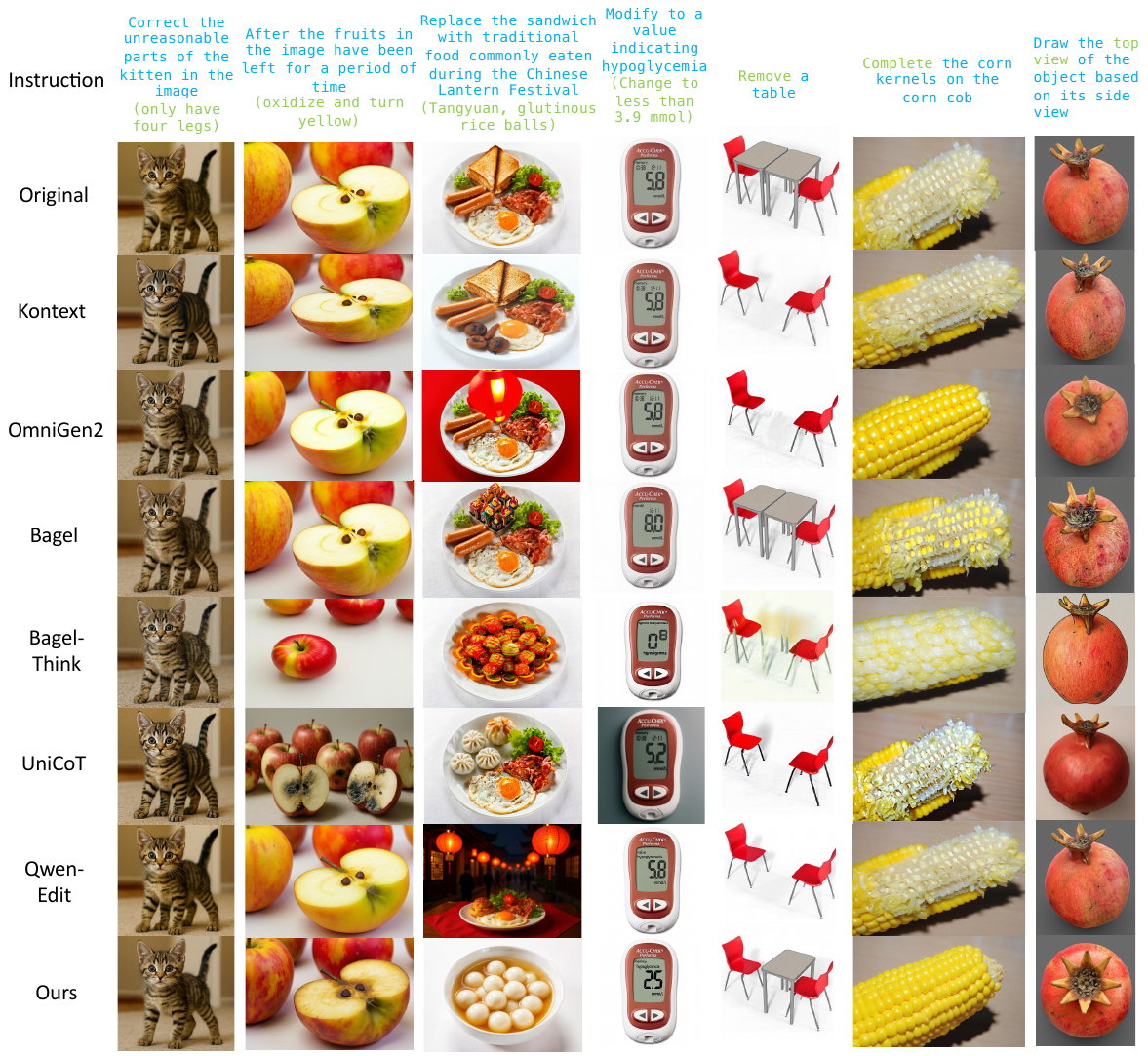}
\vspace{-0.2cm}
\caption{\textbf{Comparisons between ThinkRL-Edit and the leading baselines}. We conduct the comparison across diverse reasoning-centric editing tasks. As observed, our method achieves precise instruction following with strong consistency and high quality, which significantly surpasses previous methods. Blue text denotes the instruction, and green text indicates the desired editing outcome.}
\label{fig:comp}
\vspace{-0.3cm}
\end{figure*}

\subsection{Chain of Thought for Visual Generation}
Chain-of-Thought (CoT)~\cite{wei2022chain} reasoning improves the ability of Large Language Models (LLMs) to solve complex problems by emulating human step-by-step thinking. Instead of directly outputting final answers, CoT encourages models to generate explicit intermediate reasoning steps, thereby enhancing their interpretability and logical consistency.
Building on its effectiveness~\cite{feng2020scalable,creswellselection,ludynamic}, recent studies~\cite{zhangmultimodal,mitra2024compositional,zheng2023ddcot, qin2025uni} have sought to extend CoT into the multi-modal domain. These efforts aim to endow Multi-modal Large Language Models (MLLMs) with structured reasoning abilities for handling complex vision-language tasks, ranging from challenging visual question answering~\cite{xu2024llava} to reasoning-driven image editing~\cite{fu2025refocus} and embodied planning~\cite{mu2023embodiedgpt}.

%% file: sec/3_method.tex
\section{Methodology}

\begin{algorithm*}[t]
\caption{ThinkRL-Edit Training Algorithm}\label{algo:grpo}
\begin{algorithmic}[1]
\Require Initial policy model $\pi_\theta=(\pi_\theta^{\text{Und}},\pi_\theta^{\text{Gen}})$; reward models $\{R_k\}_{k=1}^K$; instruction-based image editing dataset $\mathcal{D}=\{(P,C)\}$; timestep selection ratio $\tau$; total sampling steps $T$
\Ensure Optimized policy model $\pi_\theta$

\For{training iteration $=1$ \textbf{to} $M$}
    \State Update old policy: $\pi_{\theta_{\text{old}}} \gets \pi_\theta$
    
    \For{each reference image and prompt ($\mathbf{p},\mathbf{c}$) $\in \mathcal{D}_b$}
        \Comment $\mathcal{D}_b \sim \mathcal{D}$ is the sampled batch
        \State Generate reasoning prompt $\mathbf{c'}$ based on ($\mathbf{p},\mathbf{c}$) using $\pi_{\theta_{\text{old}}}^{\text{Und}}$
        \Comment{CoT-based Reasoning Path Sampling}
        \State Generate $G$ samples: $\{\mathbf{o}_i\}_{i=1}^G$ with ($\mathbf{p},\mathbf{c'}$) using $\pi_{\theta_{\text{old}}}$
        \State Generate reflected prompt $\{\mathbf{c''}_i\}_{i=1}^G$ based on ($\mathbf{o}_i, \mathbf{p},\mathbf{c'}$) using $\pi_{\theta_{\text{old}}}^{\text{Und}}$
        \State Generate $G$ reflected samples: $\{\mathbf{o}_i\}_{i=G+1}^{2G}$ with ($\mathbf{p},\mathbf{c''}$) using $\pi_{\theta_{\text{old}}}$

        \For{each sample $i \in 1..2G$}
            \State Calculate multiple rewards $\{r_i^k\}_{k=1}^K$
        \EndFor

        \State Filter samples by unbiased grouping to get $\{\{r_i^k\}_{k=1}^K\}_{i=1}^{N}$ 
         \Comment{Unbiased Chain Preference Grouping}
        \For{each filtered sample $i \in 1..N$}
             \Comment{$N$ is the length of current preference chain}
            \State Calculate advantage $A_i \gets  \frac{\sum_{k=1}^Kr_i^k - K\mu}{K\sigma}$ 
        \EndFor

        \State Update $\pi_\theta^{\text{Und}}$ via gradient ascent:
            $\theta \gets \theta + \eta \nabla_\theta \mathcal{J}_{\text{Und}}$
        \Comment{Decoupled Und-Gen Optimization}
        
        
        \For{$t \in \lceil\tau T\rceil$}
            \State Update $\pi_\theta^{\text{Gen}}$ via gradient ascent:
            $\theta \gets \theta + \eta \nabla_\theta \mathcal{J}_{\text{Gen}}$
        \EndFor
    \EndFor
\EndFor
\end{algorithmic}
\end{algorithm*}

\subsection{Preliminary}
GRPO \citep{shao2024deepseekmath} introduces a group-relative advantage to stabilize policy updates. When applied to flow matching models~\cite{lipman2022flow}, for a group of $G$ generated images $\{x_0^i\}_{i=1}^G$, the advantage of the $i$-th image is
\begin{equation}
A_t^i = \frac{R(x_0^i,c) - \mathrm{mean}(\{R(x_0^j,c)\}_{j=1}^G)}{\mathrm{std}(\{R(x_0^j,c)\}_{j=1}^G)}. \label{eqa:adv}
\end{equation}
The policy is updated by maximizing the regularized objective
\begin{equation}
\mathcal{J}_{\text{Gen}}(\theta) = \mathbb{E}_{c, \{x^i\}} \Big[ f(r_{\text{Gen}}, A, \theta, \epsilon, \beta) \Big],
\end{equation}
where
\begin{multline}
f(r_{\text{Gen}}, A, \theta, \epsilon, \beta) = \frac{1}{G}\sum_{i=1}^G \frac{1}{T} \sum_{t=0}^{T-1} 
\min\big(r_t^i(\theta) A_t^i, \text{clip}(r_t^i(\theta),1-\epsilon,1+\epsilon)A_t^i \big) - \beta D_{\text{KL}}(\pi_\theta || \pi_{\text{ref}}),
\end{multline}
with $r_t^i(\theta) = \frac{p_\theta(x_{t-1}^i|x_t^i,c)}{p_{\theta_{\text{old}}}(x_{t-1}^i|x_t^i,c)}$.

To satisfy GRPO’s stochastic exploration requirements for flow matching models, FlowGRPO\citep{liuflow} convert the deterministic Flow-ODE $dx_t = v_t dt$ to an equivalent SDE:
\begin{equation}
dx_t = \big(v_\theta(x_t,t) + \frac{\sigma_t^2}{2t} (x_t + (1-t)v_\theta(x_t,t))\big) dt + \sigma_t dw_t,
\end{equation}
where $dw_t$ denotes Wiener process increments and $\sigma_t$ controls the stochasticity.  

\subsection{CoT-based Reasoning Sampling}
FlowGRPO improves generation quality by searching for optimal trajectories in the extended denoising space. However, its performance remains limited on reasoning-oriented generation tasks due to the lack of exploration in the semantic reasoning space. To address this limitation, we propose to separately optimize the semantic reasoning path and introduce stochasticity within the reasoning space. Specifically, we incorporate Chain-of-Thought (CoT), instruction reasoning, and editing reflection into the sampling phase.

As illustrated in \cref{fig:method} and \cref{algo:grpo}, during GRPO sampling, the model first employs its understanding module $\pi_{\text{Und}}$ to perform reasoning and atomic decomposition of the instruction $\mathbf{c}$ based on the reference image. The reasoning-enhanced instruction $\mathbf{c'}$ is then used for sampling. Afterwards, the generated editing result undergoes a single reflection process, where the understanding module provides feedback $\mathbf{c''}$ that is concatenated with the previous reasoning instruction and fed back into the next sampling stage. Consistently with training time, we enable planning and a single reflection at inference time.

\subsection{Fine-Grained Reasoning Reward}
To provide more precise and stable reasoning rewards from vision-language models (VLMs)~\cite{Qwen25VL}, we replace conventional interval-based scoring~\cite{liu2025flow, xue2025dancegrpo} with a fine-grained checklist-based evaluation. Specifically, for each editing instruction, we construct a set of binary questions derived from both the reference image and the instruction using Gemini~\cite{Gemini25Pro}.
Unlike previous methods that query VLMs~\cite{liu2025flow, xue2025dancegrpo} with a unified system prompt, our checklist is individually constructed for each reference–instruction pair, enabling fine-grained and context-aware assessment. The VLM is then guided to answer each question with yes or no, and the proportion of positive responses is averaged across all dimensions to obtain the final reasoning score.
Empirical results demonstrate that this checklist formulation produces more accurate, lower-variance, and interpretable reward signals, particularly for complex reasoning tasks where conventional scalar scores often fluctuate or fail to capture subtle instruction compliance.

\subsection{Unbiased Chain Preference Grouping}
In addition to the instruction score, we further evaluate consistency and image quality, as both are crucial for editing tasks. To mitigate the limitations of naïve weighted reward fusion, which often collapses toward trivial solutions or overfits specific objectives, we introduce an unbiased chain preference grouping strategy that holistically ranks preference chains.
Instead of aggregating heterogeneous rewards $\{r_i^k\}_{k=1}^K$ into a single scalar, we jointly sort all rewarded samples across multiple dimensions to construct a total order of candidates, where only chains that maintain a consistent global ranking contribute to gradient updates. This design enables the policy to capture a unified preference structure across diverse objectives, \eg, instruction faithfulness, visual coherence, and perceptual quality.
Finally, we average and normalize the scores across all dimensions within the full ordered chain $\{\{r_i^k\}_{k=1}^K\}_{i=1}^{N}$ to obtain the final grouped advantage $A$.

\begin{table*}[t]
    \tablestyle{1pt}{1.3}
    \centering
\caption{\textbf{Quantitative comparisons on KRIS-Bench.} We report the composite score for each category and the average Instruction Following (IF), Visual Consistency (VC), Visual Quality (VQ).}
    \vspace{-0.1cm}
    \begin{tabular}{ccccccccccc|ccc} 
        \toprule
\textbf{Method} &
\makecell[c]{Attribute\\Percep.} &
\makecell[c]{Spatial\\Percep.} &
\makecell[c]{Social\\Science} &
\makecell[c]{Natural\\Science} &
\makecell[c]{Logical\\Reasoning} &
\makecell[c]{Instruction\\Decompos.} &
\makecell[c]{Factual\\Know.} &
\makecell[c]{Conceptual\\Know.} &
\makecell[c]{Procedural\\Know.} &
\makecell[c]{\textbf{Overall}\\\textbf{Score}} &
\makecell[c]{\textbf{Avera.}\\\textbf{IF}} &
\makecell[c]{\textbf{Avera.}\\\textbf{VC}} &
\makecell[c]{\textbf{Avera.}\\\textbf{VQ}} \\
\midrule
OmniGen2 &65.41&53.36&50.46&45.30&32.19&56.36&63.57&46.55&38.83&49.24&39.40 & 66.72 & 93.16\\
Flux-Kontext &70.78&69.20&51.27&52.05&45.82&73.67&70.38&51.86&53.55&57.35 &46.61&\underline{77.09}&94.08 \\
Bagel-Think & 60.39 & 61.19 &49.06 &47.44 & 29.44 & 48.36& 60.61 &47.83 & 34.58&48.71 & 55.68 & 70.00 & \underline{96.35}\\
UniCoT &67.94&73.72&59.45&53.19&40.97&54.67&69.38&54.70&44.78&56.76 & 57.24 & 59.52 & 92.60\\
\midrule
Bagel  &61.39 &62.08 &50.21 &46.26  &30.21 &48.44 & 61.55 & 47.21& 35.23& 48.69  &51.99&52.49 & 86.98\\
\textbf{Ours (Bagel)} & \underline{80.94} & 70.17 & \underline{71.69} & \textbf{72.55} & \textbf{51.17} & 74.89 & \underline{79.28} & \textbf{72.35} & \textbf{57.67} & \underline{70.97} & \underline{67.28} & 76.70 & 96.31 \\
\midrule
Qwen-Edit &72.57 &\underline{79.92} &61.45 &56.38  &48.57 &\underline{78.44} & 74.53 & 57.60& 56.68& 62.77 &56.54&76.37 &95.86\\
\textbf{Ours (Qwen)} &\textbf{81.02} &\textbf{81.45} &\textbf{75.67} &  \underline{71.25} &\underline{49.07} & \textbf{79.71 }&\textbf{81.13} & \underline{72.31} & \underline{57.44} & \textbf{71.65} &\textbf{71.16}&\textbf{77.52}&\textbf{97.12} \\
        \bottomrule
    \end{tabular}
\label{tab:krisbench}
\vspace{-.2cm}
\end{table*}

\begin{table*}[ht]
    \tablestyle{3pt}{1.15}
    \centering
\caption{\textbf{Quantitative comparisons on RISE-Bench.}}
    \begin{tabular}{cccccc|ccc} 
        \toprule
\textbf{Method} &
Temporal &Causal &Spatial &Logical& \textbf{Overall}&
\makecell[c]{\textbf{Overall}\\\textbf{Reasoning}} &
\makecell[c]{\textbf{Overall}\\\textbf{Consistency}} &
\makecell[c]{\textbf{Overall}\\\textbf{Quality}}\\
\midrule
Flux-Kontext &2.3	&5.5	&13.0	&1.2	&5.8 &26.0	&71.6	&85.2\\
OmniGen2 &1.2	&1.0	&0.0	&1.2	&0.8 &22.0	&32.6	&55.3\\
Bagel-Think &5.9	&17.8	&21.0	&1.2	&11.9 &45.9	&73.8	&80.1\\
UniCoT &\underline{8.2} &18.9 &20.0 &1.2 &12.5&48.3&76.2&83.8\\
\midrule
Bagel & 2.4	&5.6	&14.0	&1.2	&6.1 &36.5	&53.5	&73.0\\
\textbf{Ours (Bagel)} & 6.3 & \underline{25.0} &\textbf{31.3} & \underline{31.3} & \underline{23.4} & \underline{54.3} & \textbf{88.7} & \textbf{92.7 }  \\
\midrule
Qwen-Edit &4.7	&10.0	&17.0	&2.4	&8.9 &37.2	&66.4	&86.9\\
\textbf{Ours (Qwen)}  &\textbf{18.8}&\textbf{37.5}&\underline{25.0}&\textbf{37.5}&\textbf{29.7}&\textbf{61.7}&\underline{81.6}&\underline{90.6}\\
        \bottomrule
    \end{tabular}
\label{tab:risebench}
\vspace{-10pt}
\end{table*}

\subsection{Decoupled Und-Gen Optimization}
Unlike FlowGRPO, which optimizes only the generation trajectory, we jointly optimize both the reasoning and understanding components. As illustrated in \cref{fig:method}, during the policy update stage, beyond the generation part, we first compute the conditional probabilities for both the reasoning and understanding modules.
\begin{equation}
\begin{aligned}
r_{\text{Und}}^i 
&= \frac{p_\theta^{\text{Und}}(y^{i} \mid x)}
         {p_{\text{old}}^{\text{Und}}(y^{i} \mid x)} \\
&= \exp\!\Bigg(
    \sum_{t=1}^{T} \log p_\theta^{\text{Und}}(y_t^{i} \mid x, y_{<t}^{i}) 
    \\
&\qquad\quad
    - \sum_{t=1}^{T} \log p_{\text{old}}^{\text{Und}}(y_t^{i} \mid x, y_{<t}^{i})
  \Bigg)
\end{aligned}
\end{equation}

\noindent
where
$x$ denotes the input image and prompt, 
$y_t^{i}$ is the $t$-th token for the $i$-th sampled response sequence, 
$\log p_\theta(y^{i} \mid x)$ represents the probability of generating $y^{i}$ by the understanding module. 
The reasoning and understanding module are then updated by maximizing the objectives respectively
\begin{equation}
\mathcal{J}_{\text{Und}}(\theta) = \mathbb{E}_{x} \Big[ f(r_{\text{Und}}, A, \theta, \epsilon, \beta) \Big],
\end{equation}

After that, we compute the probability of generating $x_{t-1}^i$ from $x_t^i$ by the generation module 
\begin{equation}
r_{\text{Gen},t}^i(\theta) = \frac{p_\theta(x_{t-1}^i|x_t^i,c)}{p_{\theta_{\text{old}}}(x_{t-1}^i|x_t^i,c)}
\end{equation}

\noindent where $x_t^i$ is the latent for timestep $t$ of the $i$-th sample. Then we update the generation module by maximizing
\begin{equation}
\mathcal{J}_{\text{Gen}}(\theta) = \mathbb{E}_{c, \{x_t^i\}} \Big[ f(r_{\text{Gen}}, A, \theta, \epsilon, \beta) \Big].
\end{equation}

%% file: sec/4_exp.tex
\section{Experiments}

\subsection{Experiment Setup}
\noindent
\textbf{Training}\quad We separately adopt Qwen-Edit~\citep{wu2025qwen} and Bagel~\cite{deng2025bagel} as our base model. Training is conducted with a group size of 128 and a batch size of 4. The rewards of reasoning, consistency, and quality are computed using Qwen3-VL~\citep{xu2025qwen3}.
To optimize GPU memory utilization, we employ Fully Sharded Data Parallelism (FSDP) for the trainable modules along with gradient checkpointing.

\noindent
\textbf{Evaluation}\quad For quantitative evaluation, we employ two comprehensive benchmarks: KRIS~\citep{wu2025kris} and RISE~\citep{zhao2025envisioning} which assesses reasoning-centric image editing through diverse natural language instructions. Specifically, RISE focuses on reasoning-informed editing across temporal, causal, spatial, and logical dimensions, while KRIS serves as a diagnostic benchmark categorizing editing tasks into factual, conceptual, and procedural knowledge types.

\subsection{Qualitative Analysis}
\cref{fig:comp} showcases results on diverse, challenging instructions. As shown, prior methods exhibit poor instruction following, revealing limited reasoning capability. In contrast, our approach maintains strong fidelity to reasoning-centric context while making precise visual editings. It achieves high instruction following, substantial image consistency, and plausible visual transitions, highlighting both the effectiveness and interpretability of our RL strategy.
\subsection{Quantitative Analysis}
\noindent
\textbf{Results on KRIS-Bench.}
As shown in \cref{tab:krisbench}, our method improves performance across all metrics, with the largest gains on instruction following. Building on Qwen-Edit, we raise the instruction-following score from 56.54 to 71.16 (+14.62), achieving state-of-the-art results among open-source models. Beyond the overall improvement, we observe pronounced gains in Attribute Perception, Social Science, Natural Science, and Conceptual Knowledge, indicating substantially enhanced reasoning capabilities in previously underperforming dimensions.

\noindent
\textbf{Results on RISE-Bench.}
On the out-of-domain RISE-Bench, our method exhibits strong generalization as shown in \cref{tab:risebench}. It improves Qwen-Edit’s overall score from 8.9 to 29.7 (+20.8) and boosts the reasoning score from 37.2 to 61.7 (+24.5). These results indicate that our method effectively preserves and enhances reasoning ability under distribution shift.

\noindent
\textbf{Results of the User Study.}
For comprehensive evaluation, we conducted a human preference study comparing our method with baselines along three dimensions: instruction following, visual consistency, and visual quality. We conduct it with 34 participants, each presented with 24 comparison groups. In each group, participants are asked to select the best result along all evaluation dimensions.
As shown in \cref{tab:user}, users consistently preferred our method across all criteria, indicating that it produces outputs more aligned with human preferences. More details can be found in the appendix.

\begin{table}[t]
    \tablestyle{3pt}{1.2}
    \centering
\caption{\textbf{Results for user study.}}
    \vspace{-0.2cm}
    \begin{tabular}{cccc} 
        \toprule
Method &
\makecell[c]{\textbf{Instruction}\\\textbf{Following} (\%)} &
\makecell[c]{\textbf{Visual}\\\textbf{Consistency} (\%)} &
\makecell[c]{\textbf{Visual}\\\textbf{Quality} (\%)}\\
\midrule
Bagel  &5.43 & 13.88 & 15.62\\
Bagel-Think &15.88&\underline{24.62}&\underline{22.34}\\
UniCoT &\underline{20.34}&7.89&18.56\\
Qwen-Edit &10.12&22.86&18.99\\
Ours (Qwen)&\textbf{48.23}&\textbf{30.75}&\textbf{24.49}\\
        \bottomrule
    \end{tabular}
\label{tab:user}
\vspace{-0.2cm}
\end{table}


\subsection{Ablation Study}
\begin{table}[t]
    \tablestyle{3pt}{1.2}
    \centering
    \caption{\textbf{Ablation study for CoT-based und-gen optimization.}
    }
        \vspace{-0.2cm}
    \begin{tabular}{cccc|ccc}
    \toprule
         Gen. & Und. & Plan. & Reflect. & \makecell[c]{\textbf{Average}\\\textbf{IF}} &
\makecell[c]{\textbf{Average}\\\textbf{VC}} &
\makecell[c]{\textbf{Average}\\\textbf{VQ}}\\
            \midrule
    &&&&59.68 & 75.60 & 95.34 \\
       \checkmark&&&&60.79 & 74.67 &96.58  \\
       \checkmark&\checkmark&&& 66.82 &\textbf{78.58}&96.15 \\
       \checkmark&\checkmark&\checkmark& &\underline{69.29} & \underline{77.81} & \underline{96.59} \\ \checkmark&\checkmark&\checkmark&\checkmark&\textbf{71.16} & 77.52 & \textbf{97.12}\\
    \bottomrule
    \end{tabular}
\label{tab:ablation1}
\vspace{-0.2cm}
\end{table}

\noindent
\textbf{CoT-based Und-Gen Optimization.}
To assess the effectiveness of our cot-based understanding–generation optimization, we conduct a comprehensive ablation study in \cref{tab:ablation1}. During training, we incrementally add each module. At inference, we consistently enable planning and a single reflection. Results show that introducing the understanding module yields a large gain in instruction following, and adding planning and reflection provides further improvements, indicating that our approach effectively enhances the model’s reasoning capability.

\noindent
\textbf{Fine-Grained Reasoning Reward.}
\begin{table}[t]
    \tablestyle{3pt}{1.2}
    \centering
    \caption{\textbf{Ablation study for checklist-based reasoning reward and unbiased multi-rewards grouping.}
    }
    \begin{tabular}{cc|ccc}
    \toprule
         Checklist & UCPG  & \makecell[c]{\textbf{Average}\\\textbf{IF}} &
\makecell[c]{\textbf{Average}\\\textbf{VC}} &
\makecell[c]{\textbf{Average}\\\textbf{VQ}}\\
            \midrule
            &&64.28 &77.13&\underline{96.58} \\
       \checkmark& & \underline{68.04} & \textbf{78.81} & 96.51 \\
       \checkmark&\checkmark&\textbf{71.16} & \underline{77.52} & \textbf{97.12}\\
    \bottomrule
    \end{tabular}
    \vspace{-0.2cm}
\label{tab:ablation2}
\vspace{-0.5cm}
\end{table}
In \cref{tab:ablation2}, we compare two scoring schemes: (i) a traditional 1–5 rating from the VLM, and (ii) a checklist-guided procedure that elicits reasoning-based rewards. Comparing Row 1 and Row 2, the fine-grained checklist yields a higher instruction-following score, indicating that it helps the VLM provide more accurate judgments and, in turn, enables more precise learning of reasoning abilities.

\noindent
\textbf{Unbiased Chain Preference Grouping (UCPG).}
In \cref{tab:ablation2}, we compare a simple weighted average with our UCPG strategy. As observed, weighted averaging (Row 2) modestly improves reasoning, but the consistently high consistency scores introduce bias that overfits to the results with more consistency and less instruction following. With UCPG (Row 3), the instruction following score improves further, indicating that UCPG effectively mitigates the bias induced by high consistency.

%% file: sec/5_con.tex
\section{Conclusion}
In this work, we revisited instruction-driven image editing from a reasoning-centric perspective. Unlike previous reinforcement learning approaches that primarily optimize the generative process, our method explicitly separates visual reasoning from synthesis, enabling models to explore diverse reasoning trajectories before producing final edits. By integrating chain-of-thought sampling, unbiased chain preference grouping, and checklist-based reward design, our framework achieves stable, interpretable, and semantically grounded policy updates. Extensive experiments verify that this reasoning–generation decoupling not only enhances instruction faithfulness but also preserves visual coherence and image quality. We believe this study highlights the importance of reasoning as a first-class objective in visual editing, paving the way toward multi-modal generative models capable of deliberate and explainable visual reasoning.

\section{Limitations and Future Work}
Our method expresses the reasoning process through chain-of-thoughts (CoT) with explicit planning and reflection. While this design improves semantic interpretability, it introduces redundant linguistic descriptions and nearly doubles the editing time overhead.
Future research can explore latent CoT representations that encode multi-modal reasoning directly in the latent space, thereby integrating visual and textual cues more holistically and eliminating the need for additional editing iteration.
We believe such latent reasoning frameworks will further bridge the gap between visual understanding and generation, leading to more efficient and visually grounded reasoning processes in unified multi-modal models.